\documentclass[journal]{IEEEtran}

\usepackage{cite}
\usepackage{amsmath,amssymb}
\usepackage{graphicx}
\usepackage[caption=false,font=footnotesize]{subfig}
\usepackage{booktabs}
\usepackage{multirow}
\usepackage{makecell}
\usepackage{bm}
\usepackage{stfloats}
\usepackage{afterpage}

\makeatletter
\renewcommand\subsection{\@startsection{subsection}{2}{\z@}%
  {1.0ex plus 0.2ex minus 0.1ex}%
  {0.6ex plus 0.1ex minus 0.1ex}%
  {\normalfont\normalsize\itshape}}
\renewcommand\subsubsection{\@startsection{subsubsection}{3}{\z@}%
  {0.8ex plus 0.2ex minus 0.1ex}%
  {0.5ex plus 0.1ex minus 0.1ex}%
  {\normalfont\normalsize\itshape}}
\makeatother

\begin{document}

\title{Mobile UMI: Cross-View Diffusion Policy with\\Decoupled Kinematics for Mobile Manipulation}

\author{Haoran~Huang$^\dagger$,
        Haonan~Dong$^\dagger$,
        and~Huixu~Dong%
\thanks{Manuscript received XX, 2026; revised XX, 2026.}%
\thanks{Haoran Huang and Haonan Dong contributed equally to this work.}%
\thanks{Huixu Dong is the corresponding author.}%
\thanks{The authors are with Zhejiang University, Hangzhou 310027, China (e-mail: xxx@zju.edu.cn).}}

\maketitle

\begin{abstract}
Mobile imitation learning on portable demonstration interfaces faces two coupled bottlenecks: locomotion-contaminated action labels and inference-induced execution latency on a continuously moving base. Recent wrist-mounted interfaces lower the cost of tabletop data collection, yet a single wrist view does not capture the global context required for base navigation. Adding a body-mounted camera entangles human walking with hand motion. Meanwhile, generative policies introduce hundreds of milliseconds of inference latency, during which the base advances past the predicted waypoints, forcing backward corrections at every action splice. This paper presents Mobile UMI, a hardware-free demonstration framework that addresses both gaps through three components. First, a dual-camera capture system records chest-centric global context and wrist-centric local interaction without any robot present. Second, a one-shot ChArUco-based spatial anchor unifies the chest and hand visual-inertial frames; the hand pose is then re-expressed relative to the chest to extract decoupled $\mathrm{SE}(3)$ manipulation and $\mathrm{SE}(2)$ base trajectories. Third, an asynchronous receding-horizon executor performs online state matching: each newly generated action chunk is realigned with the current physical pose so that expired waypoints are discarded before execution. The full system is evaluated on four long-horizon household tasks, attaining an average success rate of 83.8\% over 100 trials per task. Controlled comparisons against ACT and Diffusion Policy show that the chest-relative label alone closes a large portion of the gap; the online state matching mechanism closes the remainder. The results indicate that, for mobile imitation learning under the tested conditions, explicit kinematic factorization combined with state-level latency alignment provides an effective solution without requiring architectural changes to the underlying policy class.

\end{abstract}


\IEEEpeerreviewmaketitle

\begin{figure*}[!t]
\centering
\includegraphics[width=\textwidth]{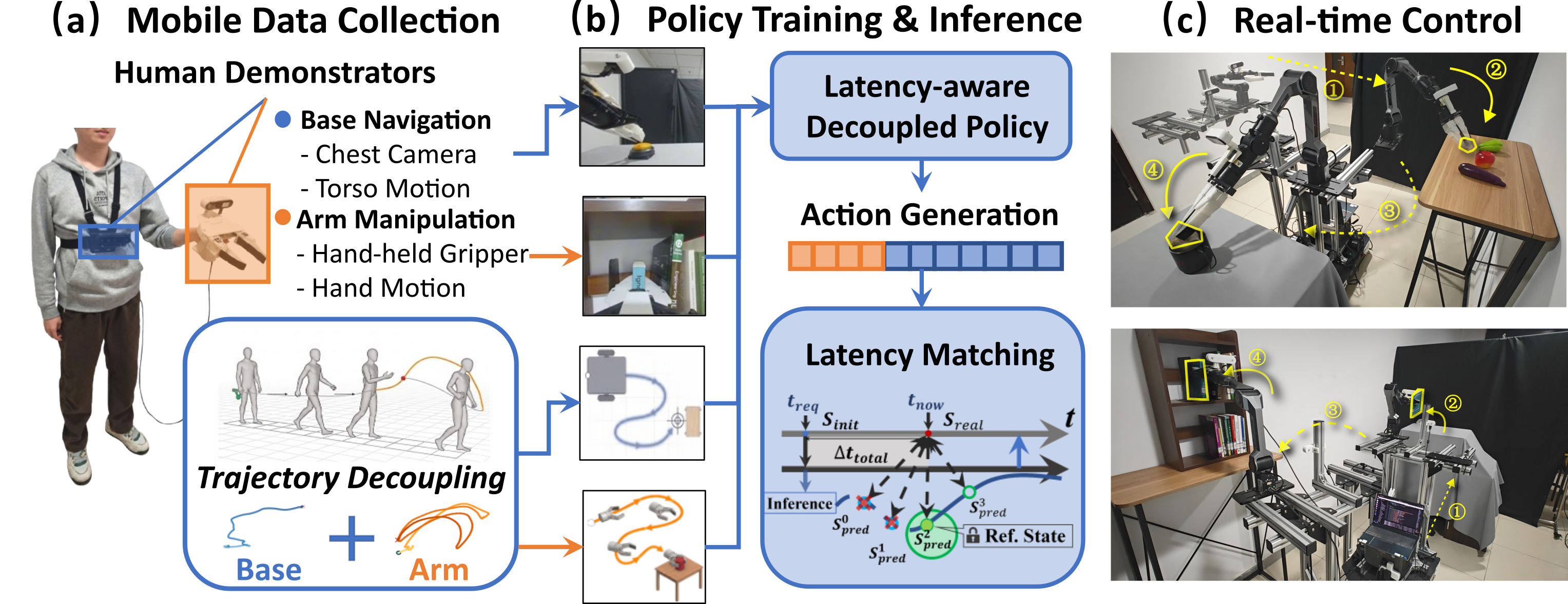}
\caption{Overview of the Mobile UMI framework.
Left: a human operator wearing a chest camera and carrying a handheld camera-gripper module collects demonstrations freely, without any robot present (\textit{Robot-Free \& Untethered}); the dual-camera design simultaneously captures global navigation context and local manipulation detail.
Center: an offline processing pipeline applies visual-inertial odometry (VIO), spatial anchoring via a static ChArUco board, and a relative-pose transformation to extract decoupled $\mathrm{SE}(2)$ base navigation trajectories and relative $\mathrm{SE}(3)$ hand manipulation trajectories, which are used to train a cross-view conditional diffusion policy.
Right: the learned policy is deployed on a differential-drive mobile manipulator; an asynchronous receding-horizon executor with spatial-temporal state matching compensates for inference latency, enabling smooth closed-loop execution without trajectory rollback or mechanical jitter.}
\label{fig:overview}
\end{figure*}

\section{Introduction}
\IEEEPARstart{S}{ervice} robots are transitioning from structured industrial settings to dynamic, unstructured environments such as homes. Mobile manipulation in these environments requires tight coordination between base navigation and dexterous arm interaction, a coupled problem that manual programming struggles to solve. Imitation learning from human demonstrations has therefore emerged as a promising paradigm. This paper proposes Mobile UMI, a hardware-free demonstration framework that addresses two coupled bottlenecks of mobile imitation learning: kinematic entanglement between the human body and the hand, as well as execution latency of generative policies on a continuously moving base.

Obtaining high-quality demonstrations on mobile platforms remains a bottleneck. Conventional teleoperation is expensive, requires the physical robot, yet remains computationally unintuitive for operators. Recent portable interfaces, such as the Universal Manipulation Interface (UMI)~\cite{b1}, lower this cost on static tabletop tasks by relying on a wrist-mounted camera. Extending this paradigm to mobile manipulation, however, exposes two limitations. A single wrist view does not capture the global spatial context required for large-scale base navigation. Adding a chest-mounted camera introduces a new problem: every operator step shifts the global hand pose by the same amount as the torso. A policy trained on these coupled signals cannot distinguish intentional arm motion from walking-induced translation.

Beyond representation challenges, generative policies introduce severe execution issues on a moving base. Algorithms that model multimodal distributions, such as diffusion policy~\cite{b2}, often incur hundreds of milliseconds of inference latency. For a stationary arm this delay is tolerable; for a continuously moving base, the robot advances during inference. The first waypoint of the returned trajectory therefore lies behind the current pose. Executing it forces a backward correction that manifests as visible mechanical jitter and trajectory rollback.

To overcome these intertwined spatial and temporal challenges, Mobile UMI integrates three components, as shown in Figure~\ref{fig:overview}. A portable dual-camera system captures chest-centric global context and wrist-centric local interaction. A one-shot ChArUco-based spatial anchor unifies the two visual-inertial frames; the hand pose is then re-expressed relative to the chest to extract decoupled $\mathrm{SE}(3)$ manipulation and $\mathrm{SE}(2)$ base trajectories. A cross-view conditional diffusion policy is trained on these decoupled labels. An asynchronous receding-horizon executor performs online state matching at each splice, realigning the newly generated chunk with the current physical pose so that expired waypoints are discarded before execution.

The principal contributions are summarized as follows:
\begin{itemize}
\item A decoupled kinematic representation that explicitly separates human locomotion from local hand interaction, enabling neural policies to learn mobile manipulation efficiently.
\item An online spatial-temporal state matching mechanism that compensates for inference delay, ensuring smooth closed-loop execution on a continuously moving platform.
\item Extensive real-world experiments on four long-horizon household tasks, in which Mobile UMI substantially outperforms ACT and Diffusion Policy baselines under matched action representations.
\end{itemize}

\section{Related Work}
\label{sec:related_work}

This section reviews prior work across three interconnected areas: mobile manipulation platforms and teleoperation systems, portable and egocentric demonstration interfaces, and execution latency in generative visuomotor policies. For each area, we identify the core limitation that our approach addresses. Early work established the feasibility of extracting manipulation priors from human videos or static datasets~\cite{b3,b4,b5,b6}, yet these methods provided no mechanism for data collection on mobile platforms.

\subsection{Mobile Manipulation and Teleoperation Platforms}
\label{sec:rw_mobile}

Acquiring high-fidelity demonstration data for mobile manipulation traditionally relies on integrated hardware systems that require the physical presence of the robot. Wu \emph{et al.} introduced TidyBot++, an open-source holonomic mobile manipulator for learning-based household tasks~\cite{b7}. Sundaresan \emph{et al.} proposed HoMeR, extracting hybrid imitation priors via a whole-body controller~\cite{b8}. Mei \emph{et al.} designed MobRT, leveraging a digital twin framework to scale data collection~\cite{b9}. Bimanual mobile systems extend these platforms with a second arm~\cite{b10,b11,b12,b13,b14}. Three limitations constrain their scalability. First, every demonstration requires the physical robot to be present, preventing data collection in novel environments without logistics overhead. Second, operators must master specialized controllers or VR headsets, imposing a skill barrier. Third, kinematics learned on one embodiment do not transfer directly to morphologically different robots. The hardware-free demonstration paradigm introduced in this paper removes all three constraints.

\subsection{Portable and Egocentric Demonstration Interfaces}
\label{sec:rw_portable}

To eliminate the constraints of whole-body teleoperation, researchers turned to wearable and handheld demonstration interfaces. Chi \emph{et al.} developed the Universal Manipulation Interface (UMI), which collects data through a wrist-mounted camera affixed to a portable gripper~\cite{b1}. Wang \emph{et al.} designed DexCap using motion-capture gloves~\cite{b15}; Xu \emph{et al.} introduced DexUMI using a tactile exoskeleton~\cite{b16}. A range of related systems subsequently emerged~\cite{b17,b18,b19,b20,b21,b22,b23,b24,b25,b26}. These portable interfaces decouple data acquisition from the robot yet scale well only on static tabletop tasks. Extending them to mobile manipulation exposes a geometric conflict. A single wrist-mounted camera lacks the field of view needed for meter-scale base navigation. Adding a chest-mounted camera introduces a new problem: every operator step shifts the global hand pose by the same amount as the torso displacement. A policy can no longer distinguish intentional arm motion from walking-induced translation. The framework in this paper addresses this gap by computing hand poses relative to the chest frame, removing the locomotion component before learning.

\subsection{Execution Latency in Generative Policies}
\label{sec:rw_latency}

Recent visuomotor architectures adopt generative models to characterize multimodal human behaviors. The Diffusion Policy formulated by Chi \emph{et al.} mitigated the averaging effect inherent to deterministic regression~\cite{b2}. Researchers scaled action generation across 3D representations and diverse transformer backbones~\cite{b27,b28,b29,b30,b31,b32,b33}. These models capture complex action distributions with high fidelity, yet iterative denoising imposes end-to-end inference delays that routinely exceed 100ms. On a stationary arm, such a delay is tolerable. On a continuously translating mobile base, the robot may travel several centimeters during a single inference pass; the first waypoint of the returned trajectory is therefore already behind the robot's current position, forcing a backward correction that appears as visible jitter. Zhao \emph{et al.} partially alleviated command-level latency through receding horizon execution with overlapping action chunks~\cite{b27}, yet that approach lacks a mechanism to realign the stale sequence with the current physical state at each splice. The state matching mechanism proposed in this paper closes the latency gap in a model-free manner: by querying the real-time pose from VIO~\cite{b34} and identifying the nearest-neighbor waypoint on the predicted trajectory, the system discards all expired commands before resuming execution, eliminating jitter without requiring explicit dynamics models or constraint solvers.

\section{Data Collection System}
\label{sec:method_data}

Converting raw multi-camera observations into a learnable representation for mobile manipulation requires a systematic pipeline spanning hardware design, pose estimation, spatial alignment, and action decoupling. This section describes each stage.

\subsection{Hardware Architecture}
\label{sec:hardware}

The platform separates the demonstration collection end from the robot execution end, as illustrated in Fig.~2. During demonstrations, a human operator wears a chest-mounted camera and carries a handheld camera-gripper module; no robot is present. After offline processing and policy training, the learned model is transferred to the execution end.

\afterpage{%
\begin{figure}[!t]
\centering
\includegraphics[width=\columnwidth]{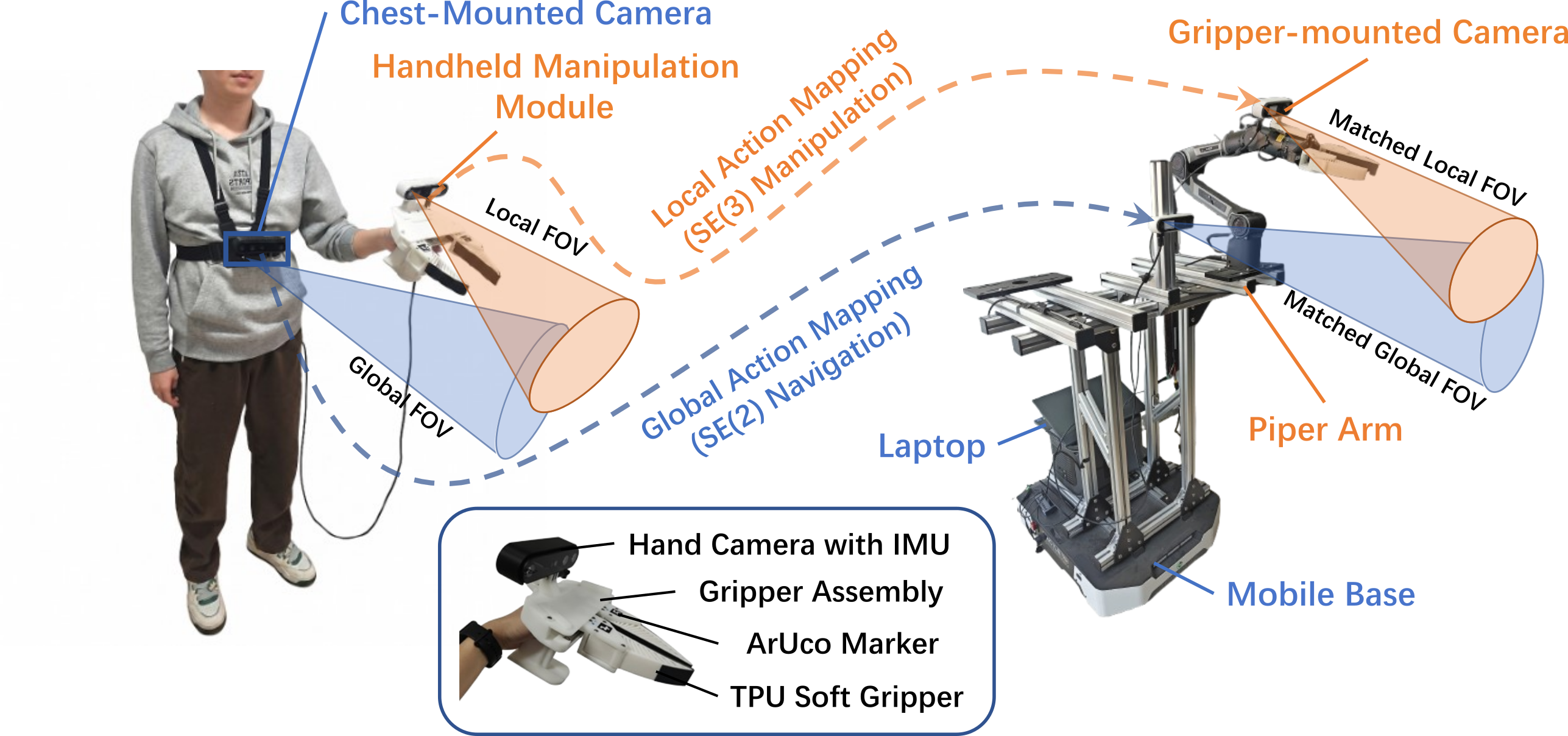}
\caption{Dual-view demonstration end and embodiment mapping to the execution robot.
Blue denotes global navigation: the chest-mounted camera provides a wide \emph{Global FOV}, which maps to the body-mounted camera on the mobile base for \emph{Matched Global FOV}.
Orange denotes local manipulation: the handheld camera--gripper module provides \emph{Local FOV}, which maps to the gripper-mounted camera for \emph{Matched Local FOV}.
Dashed arrows indicate control correspondence: $\mathrm{SE}(2)$ for base motion, $\mathrm{SE}(3)$ for arm motion.
Inset: handheld module with hand camera/IMU, gripper frame, ArUco marker, and soft fingertips.}
\label{fig:hardware}
\end{figure}
}


\subsection{Hardware Components}
\label{sec:hardware_components}
Both the chest and hand modules employ an Orbbec Gemini~2L camera, which integrates an RGB sensor (1280$\times$800 at 30Hz) with a six-axis IMU (200Hz) under a unified hardware clock. The chest camera is secured to the operator's torso via a sport harness. The hand module rigidly attaches the same camera model to a 3D-printed parallel-jaw gripper. Symmetric ArUco markers on the two fingertips enable vision-based measurement of gripper aperture.

\textbf{Execution end.}
The execution end consists of an AgileX Tracer~2.0 differential-drive base, an AgileX Piper six-degree-of-freedom arm (626.75mm reach, $\pm$0.01mm repeatability), and a parallel gripper matched to the demonstration-end tool. A body-mounted RGB camera replicates the chest-view perspective during deployment. All modules communicate via CAN bus and are synchronized through ROS.

Because the two ends share no real-time link, demonstration data can be collected across diverse scenes without transporting the robot. Mobile UMI is the only paradigm that is robot-free, supports mobile base operation, and provides explicit kinematic decoupling, all at low cost.

\subsection{Multi-Camera Pose Estimation and Spatial Anchoring}
\label{sec:anchoring}

Each camera-IMU node independently runs a filter-based visual-inertial odometry (VIO) pipeline, OpenVINS~\cite{b34}, producing an IMU pose trajectory
\begin{equation}
\label{eq:vio_traj}
\mathcal{T}_i = \bigl\{ \bm{T}^{W_i}_{I_i}(t) \in \mathrm{SE}(3) \bigr\}_{t},
\end{equation}
where $W_i$ denotes the local world frame initialized by the $i$-th node. Because these world frames are established independently, they do not share a common origin.

To unify them, a ChArUco calibration board is placed statically in the demonstration area. At each frame where the board is visible, its pose in the camera frame $\bm{T}^{C_i}_{\mathrm{tag}}(t)$ is detected via OpenCV's ArUco detection. We adopt the convention that $\bm{T}^{A}_{B}$ denotes the transformation mapping coordinates from frame $B$ to frame $A$. Combining the camera-IMU extrinsic $\bm{T}^{C_i}_{I_i}$ with the VIO output yields the board pose in the local world frame:
\begin{equation}
\label{eq:anchor}
\bm{T}^{W_i}_{\mathrm{tag}}(t) = \bm{T}^{W_i}_{I_i}(t)\;\bm{T}^{I_i}_{C_i}\;\bm{T}^{C_i}_{\mathrm{tag}}(t).
\end{equation}
A least-squares average over all valid detections produces a robust estimate $\hat{\bm{T}}^{W_i}_{\mathrm{tag}}$. With this anchor established for both nodes, the inter-node transform follows directly:
\begin{equation}
\label{eq:cross_node}
\bm{T}^{W_c}_{W_h} = \hat{\bm{T}}^{W_c}_{\mathrm{tag}}\;\bigl(\hat{\bm{T}}^{W_h}_{\mathrm{tag}}\bigr)^{-1}.
\end{equation}

\subsection{Decoupled Action Representation}
\label{sec:decouple}

During mobile demonstrations, human locomotion and hand manipulation are geometrically entangled: every step shifts both the chest and the hand in the global frame. To disentangle these two motion components, the hand trajectory is re-expressed relative to the chest reference. Denoting the chest and hand camera poses in a common frame as $\bm{T}^{W_c}_{C_c}(t)$ and $\bm{T}^{W_c}_{C_h}(t)$, the relative hand pose is
\begin{equation}
\label{eq:relative}
\bm{T}^{C_c}_{C_h}(t) = \bigl(\bm{T}^{W_c}_{C_c}(t)\bigr)^{-1}\;\bm{T}^{W_c}_{C_h}(t).
\end{equation}
This transform cancels the shared locomotion component, retaining only the local arm motion relative to the torso.

Based on this decomposition, the policy learns three semantically distinct action branches:
\begin{itemize}
\item \textbf{Base navigation}: the $\mathrm{SE}(2)$ projection of the chest trajectory onto the ground plane, encoding planar translation and heading.
\item \textbf{Arm manipulation}: the incremental $\mathrm{SE}(3)$ change of the relative hand pose in~\eqref{eq:relative}, encoding wrist motion decoupled from walking.
\item \textbf{Gripper control}: a scalar aperture signal derived from the Euclidean distance between the two ArUco fingertip markers, linearly mapped to $[0,1]$ via a one-time open-close calibration.
\end{itemize}

\subsection{Mapping: Holonomic Gait to Non-Holonomic Base}
\label{sec:embodiment}

The demonstrated $\mathrm{SE}(2)$ chest trajectory is holonomic, whereas the execution-end differential-drive base is non-holonomic. Before training, the chest trajectory is projected onto the feasible motion submanifold:
\begin{equation}
\label{eq:nonholo}
v_t = \dot{x}_t\cos\theta_t + \dot{y}_t\sin\theta_t, \qquad \omega_t = \dot{\theta}_t,
\end{equation}
discarding the lateral component $v^{\perp}_t = -\dot{x}_t\sin\theta_t + \dot{y}_t\cos\theta_t$. To prevent drift, the data collection protocol restricts the operator to forward, backward, and in-place turning. The empirical 0.99-quantile $q_{0.99}(|v^{\perp}_t|)$ across all demonstrations is below 3cm/s.

In deployment, the low-level base controller applies a saturation filter: any lateral velocity command $v^{\perp}$ exceeding 5cm/s is clipped and smoothed via a first-order low-pass filter with a 0.2s time constant.

\afterpage{%
\begin{figure}[!t]
\centering
\includegraphics[width=\columnwidth]{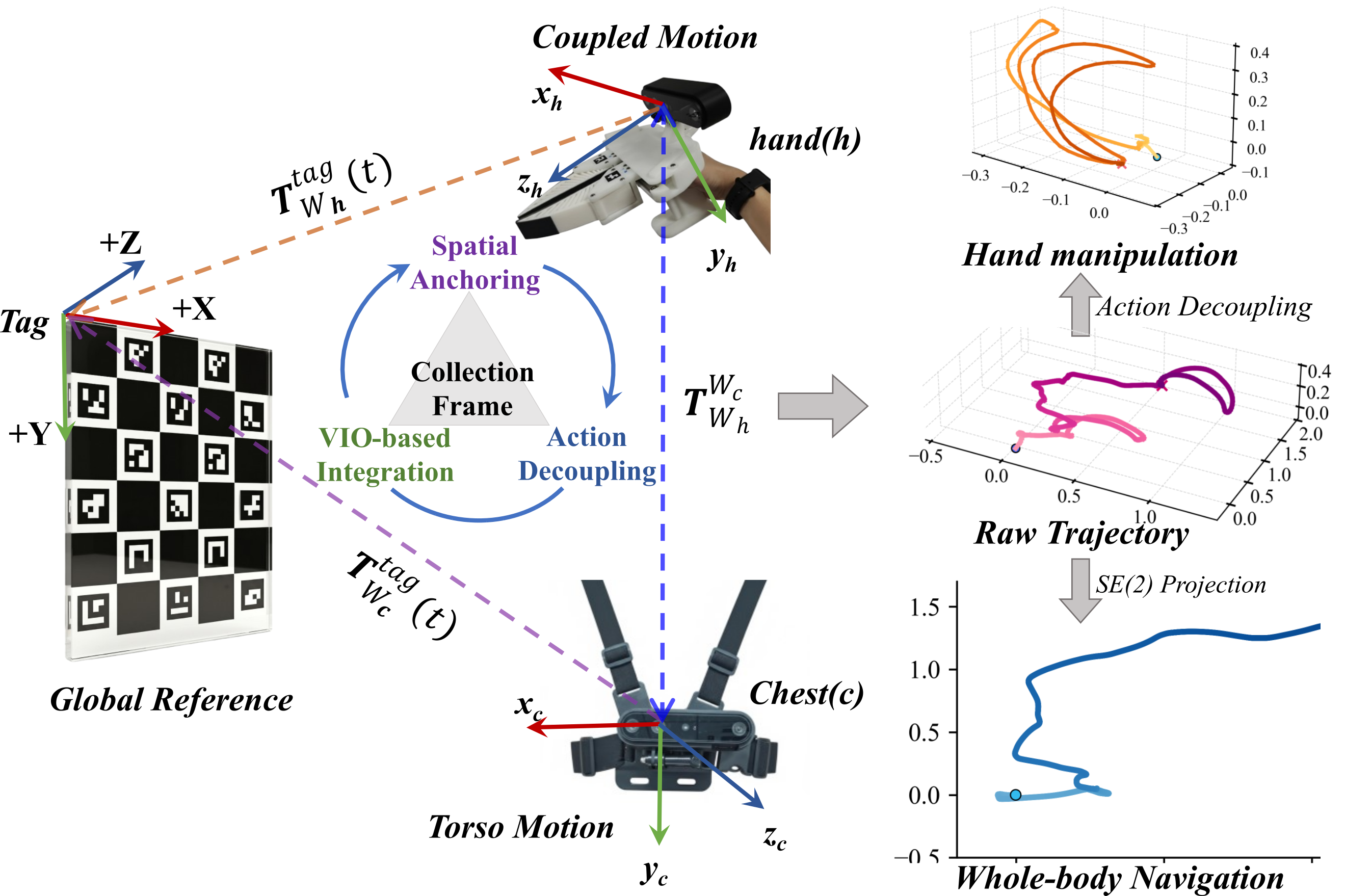}
\caption{Spatial anchoring, VIO integration, and action decoupling.
Left: a static fiducial board defines a common \emph{Tag} frame; chest ($c$) and hand ($h$) sensors each carry a local frame, with time-varying transforms $\bm{T}^{W_h}_{\mathrm{tag}}(t)$, $\bm{T}^{W_c}_{\mathrm{tag}}(t)$, and relative $\bm{T}^{W_c}_{W_h}$ linking coupled hand motion to the torso.
The circular workflow highlights spatial anchoring, VIO-based integration, and decoupling in the collection frame.
Right: the raw hand trajectory in 3D is split by \emph{Action Decoupling} into relative \emph{Hand manipulation} (isolated wrist motion) and by $\mathrm{SE}(2)$ projection into \emph{Whole-body Navigation} on the ground plane, matching the supervision branches used in Section~\ref{sec:decouple}.}
\label{fig:decouple}
\end{figure}
}

\subsection{Multimodal Demonstration Assembly}
\label{sec:assembly}

Raw trajectories from VIO, visual detection, and image streams are sampled at heterogeneous rates. A unified pipeline resamples all modalities to 10Hz: images are assigned via nearest-neighbor lookup; positions are linearly interpolated; orientations are interpolated using spherical linear interpolation (Slerp). The maximum 16.6ms synchronization latency from nearest-neighbor matching yields sub-centimeter spatial discrepancies, which are sufficiently bounded to be implicitly absorbed as observation noise. Each demonstration is quality-filtered by discarding episodes whose VIO position covariance exceeds a threshold or whose total displacement falls outside the reachable workspace. A second-order Savitzky-Golay filter smooths the pose sequences before action-label construction. The resulting dataset stores, for every 0.1s time step, the chest RGB image, hand RGB image, chest $\mathrm{SE}(2)$ pose, relative hand $\mathrm{SE}(3)$ pose, and gripper aperture.

\section{Cross-View Diffusion Policy with Latency-Aware Execution}
\label{sec:method_policy}
\suppressfloats[t]

With decoupled demonstrations in hand, two challenges remain before deployment. First, the policy must map multi-view observations to actions while preserving multimodality. Second, diffusion-based inference introduces latency that, on a continuously moving base, causes the robot to execute stale commands. This section addresses both challenges through a conditional diffusion architecture paired with a spatial-temporal delay compensation mechanism.

\subsection{Observation and Action Spaces}
\label{sec:obs_act}

At every control step $t$, the observation aggregates visual and proprioceptive signals:
\begin{equation}
\label{eq:obs}
\bm{o}_t = \bigl(\bm{I}^{c}_t,\;\bm{I}^{h}_t,\;\bm{s}_t,\;\Delta\bm{s}_{t-1}\bigr),
\end{equation}
where $\bm{I}^{c}_t$ and $\bm{I}^{h}_t$ are the chest and hand RGB images, $\bm{s}_t$ stacks the chest $\mathrm{SE}(2)$ pose, relative hand $\mathrm{SE}(3)$ pose, and gripper aperture, and $\Delta\bm{s}_{t-1}$ provides one-step historical motion context.

The action vector at step $t$ concatenates three branches into an 11-dimensional signal:
\begin{equation}
\label{eq:action}
\bm{a}_t = \bigl(
  \underbrace{\Delta x,\;\Delta y,\;\Delta\theta}_{\text{base }(3)},\;\;
  \underbrace{\Delta\bm{p},\;\Delta\bm{q}}_{\text{arm }(7)},\;\;
  \underbrace{g}_{\text{grip }(1)}
\bigr)\in\mathbb{R}^{11}.
\end{equation}
Here $\Delta\bm{p}\!=\!(\Delta p_x,\Delta p_y,\Delta p_z)\in\mathbb{R}^3$ is the translational increment, and $\Delta\bm{q}\!=\!(\Delta q_w,\Delta q_x,\Delta q_y,\Delta q_z)\in\mathbb{R}^4$ is the unit quaternion representing the rotational increment of the relative hand pose, regressed under the constraint $\|\Delta\bm{q}\|{=}1$ with a fixed hemisphere convention ($\Delta q_w\!\ge\!0$) to remove the double cover; the next-step rotation is recovered as $\bm{q}_{t+1}=\Delta\bm{q}\otimes\bm{q}_t$. The policy predicts a horizon of $T_p{=}16$ future actions in a single forward pass; only the first $T_a{=}8$ steps are dispatched for execution before re-planning. Although integrating pure increments typically suffers from compounding errors, our asynchronous state matching (Section~\ref{sec:delay}) implicitly resets the integration baseline at every chunk, effectively mitigating long-horizon drift.

\subsection{Conditional Diffusion Architecture}
\label{sec:diffusion}

Deterministic regression collapses multimodal demonstrations into their mean, producing physically invalid actions at distribution boundaries. A conditional diffusion model avoids this failure by learning the score of the action distribution and sampling via iterative denoising.

\textbf{Forward process.}
Starting from a ground-truth action chunk $\bm{a}^{0}$, Gaussian noise is incrementally added over $K{=}100$ steps according to a cosine-squared variance schedule $\bar{\alpha}_k$, yielding a noisy sample $\bm{a}^{k}=\sqrt{\bar{\alpha}_k}\,\bm{a}^{0}+\sqrt{1-\bar{\alpha}_k}\,\bm{\epsilon}$, where $\bm{\epsilon}\sim\mathcal{N}(\bm{0},\bm{I})$.

\textbf{Reverse process.}
A noise-prediction network $\bm{\epsilon}_\theta(\bm{a}^{k},k,\bm{o}_t)$ is trained to minimize
\begin{equation}
\label{eq:loss}
\mathcal{L}(\theta) = \mathbb{E}_{k,\bm{\epsilon},\bm{a}^{0}}\bigl[\|\bm{\epsilon}-\bm{\epsilon}_\theta(\bm{a}^{k},k,\bm{o}_t)\|^{2}\bigr].
\end{equation}

\textbf{Visual encoder.}
Both camera streams pass through a weight-shared ResNet-18 backbone (without the final fully connected layer), followed by spatial global average pooling. The resulting feature vectors are concatenated with the proprioceptive state $\bm{s}_t$ to form a global condition vector.

\textbf{Conditioning mechanism.}
The diffusion step $k$ is projected via sinusoidal positional encoding and a two-layer MLP into a time embedding. This embedding is concatenated with the global condition vector and fed into a Feature-wise Linear Modulation (FiLM) layer, which produces per-channel scale and shift parameters $(\gamma,\beta)$ applied at every residual block of the denoising backbone.

\textbf{Denoising backbone.}
A 1D temporal U-Net treats the action horizon as a one-dimensional sequence. The encoder path applies stride-2 convolutions to progressively compress the temporal resolution while doubling channel width; the decoder path uses transposed convolutions with skip connections to restore full resolution. The final output predicts the noise component $\hat{\bm{\epsilon}}$ of the same shape as $\bm{a}^{k}$.

At inference time, DDIM sampling reduces the required denoising iterations from 100 to 10, reducing latency below 100ms on an RTX~4070 GPU. An exponential moving average (EMA) of the network parameters with a decay factor of 0.9999 is maintained during training and used for evaluation.

\afterpage{%
\begin{figure}[!t]
\centering
\includegraphics[width=\columnwidth]{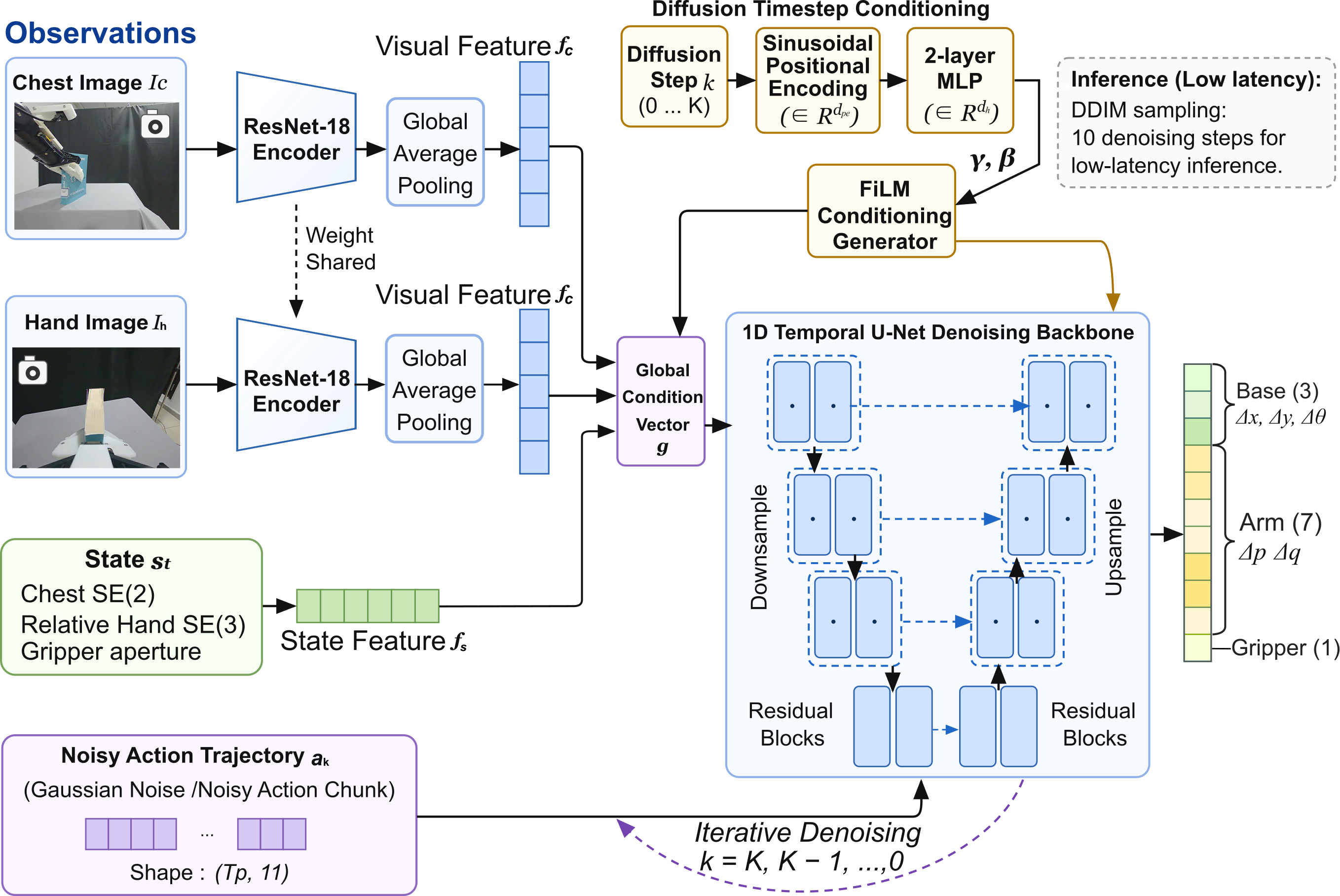}
\caption{Cross-view conditional diffusion policy architecture.
Chest and hand RGB inputs are encoded by a weight-shared ResNet-18 with global average pooling; visual features are fused with the proprioceptive state $\bm{s}_t$ into a global conditioning vector that feeds the denoising backbone.
The diffusion step $k$ is embedded sinusoidally, passed through a two-layer MLP, and injected via Feature-wise Linear Modulation (FiLM) into every residual block of the 1D temporal U-Net, which maps the noisy action trajectory $\bm{a}^{k}\!\in\!\mathbb{R}^{T_p\times 11}$ toward the clean 11-D controls ($T_p{=}16$).
At inference, DDIM sampling uses 10 steps for sub-100ms latency.}
\label{fig:network}
\end{figure}
}

\subsection{Asynchronous Receding Horizon Execution}
\label{sec:receding}

Dispatching actions synchronously with inference would stall the robot during each denoising pass. Instead, the system operates asynchronously: while the current action chunk is being executed at 10Hz, a new chunk is generated in the background conditioned on the latest observation. Upon completion, the new chunk seamlessly takes over. Because the prediction horizon ($T_p{=}16$) is twice the execution horizon ($T_a{=}8$), the overlap provides temporal continuity at the splice boundary.

\subsection{Spatial-Temporal Delay Compensation}
\label{sec:delay}

\begin{figure*}[!t]
\centering
\includegraphics[width=\textwidth]{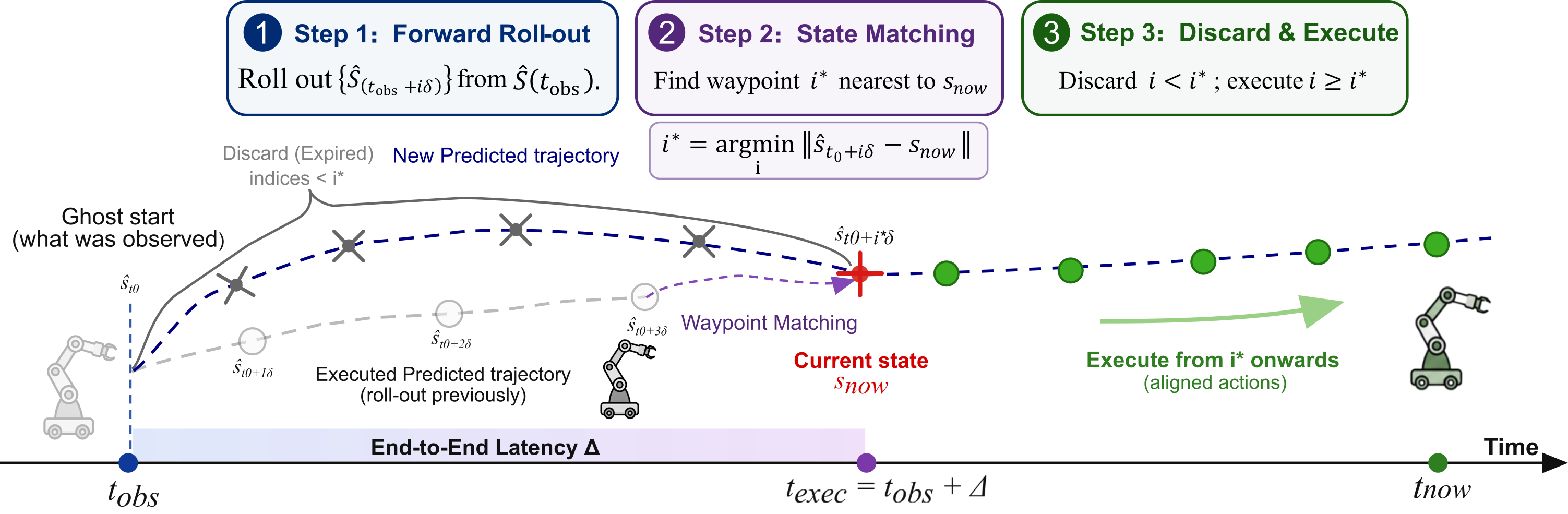}
\caption{Spatial-temporal delay compensation (asynchronous state matching).
\textbf{Step~1:} forward roll-out of predicted states $\{\hat{\bm{s}}_{t_{\mathrm{obs}}+i\delta}\}$ from the observation-time estimate $\hat{\bm{s}}(t_{\mathrm{obs}})$.
\textbf{Step~2:} weighted state-space alignment as in~\eqref{eq:match}, selecting $i^{*}$ that minimizes the weighted discrepancy between the predicted state $\hat{\bm{s}}_{t_{0}+i\delta}$ and the current measured state $\bm{s}_{\mathrm{now}}$ (odometry and proprioception).
\textbf{Step~3:} discard expired waypoints with $i<i^{*}$ and execute from $i^{*}$ onward to avoid trajectory rollback and jitter under end-to-end latency~$\Delta$.}
\label{fig:delay}
\end{figure*}

Even with DDIM acceleration, end-to-end latency $\Delta{=}\Delta_{\mathrm{in}}+\Delta_{\mathrm{net}}+\Delta_{\mathrm{exe}}$ accumulates from image capture, network inference, and motor response. Measured across 100 deployment episodes, the mean total latency is $\Delta{=}142$ms (std. 18ms), decomposed into image acquisition $\Delta_{\mathrm{in}}{=}33$ms, network inference $\Delta_{\mathrm{net}}{=}87$ms, and motor command dispatch $\Delta_{\mathrm{exe}}{=}22$ms. For a stationary arm this delay is tolerable. For a moving base traveling at 0.3m/s, the robot advances approximately 4.3cm during inference, so the first predicted waypoint already lies behind the current pose. Executing it directly forces a backward correction and produces visible jitter. Figure~\ref{fig:delay} illustrates the spatial-temporal matching logic used at deployment time.

The compensation proceeds in three steps. First, upon receiving a new action sequence, the controller forward-integrates from the observation-time state $\hat{\bm{s}}_{t_0}$ to reconstruct a predicted state trajectory $\{\hat{\bm{s}}_{t_0+i\delta}\}_{i=1}^{T_p}$. This forward roll-out is a purely \emph{kinematic} operation: it applies the predicted velocity commands to the geometric state without modeling physical inertia, motor response lag, or acceleration limits. Each predicted state decomposes into a base pose $\hat{\bm{b}}_{i}\!\in\!\mathrm{SE}(2)$, a chest-relative hand position $\hat{\bm{p}}_{i}\!\in\!\mathbb{R}^{3}$, a chest-relative hand orientation $\hat{\bm{R}}_{i}\!\in\!\mathrm{SO}(3)$, and a scalar gripper aperture $\hat{g}_{i}\!\in\!\mathbb{R}$. Second, the current physical state is queried from encoders and odometry and decomposed analogously into $(\bm{b}_{\mathrm{now}},\bm{p}_{\mathrm{now}},\bm{R}_{\mathrm{now}},g_{\mathrm{now}})$. The discrepancy between the kinematic prediction and the actual measured state arises from the \emph{dynamics} of the physical system—specifically, the base's inertia and motor response characteristics—which are not explicitly modeled in the forward roll-out. Third, the index $i^{*}$ that minimizes the weighted state discrepancy between the predicted and current states is identified:
\begin{equation}
\label{eq:match}
\begin{aligned}
i^{*}=\arg\min_{i}\Bigl(
&w_{b}\,d_{\mathrm{SE}(2)}^{\,2}(\hat{\bm{b}}_{i},\bm{b}_{\mathrm{now}})
 + w_{t}\,\|\hat{\bm{p}}_{i}-\bm{p}_{\mathrm{now}}\|^{2} \\
&+ w_{r}\,d_{\mathrm{SO}(3)}^{\,2}(\hat{\bm{R}}_{i},\bm{R}_{\mathrm{now}}) \\
&+ w_{g}\,(\hat{g}_{i}-g_{\mathrm{now}})^{2}
\Bigr).
\end{aligned}
\end{equation}
where the four error terms correspond to base $\mathrm{SE}(2)$ discrepancy, arm translational error, arm rotational geodesic distance on $\mathrm{SO}(3)$, and gripper-aperture deviation. The base term $d_{\mathrm{SE}(2)}$ folds the heading into a length quantity by scaling it with a 0.5m turning radius; $d_{\mathrm{SO}(3)}(\hat{\bm{R}},\bm{R})=\|\log(\hat{\bm{R}}^{\top}\bm{R})\|$ returns the geodesic angle. The weights are set to $(w_b,w_t,w_r,w_g){=}(1.0,1.0,0.2,0.1)$ to balance translational and rotational units while assigning lower priority to orientation and gripper state. A sweep over $[0.5\times,2\times]$ around these values changes the average task success rate by less than $\pm$2 pp, indicating robustness to the exact weighting.

Across 400 deployment trials, the selected index $i^{*}$ has a mean of 3.2 (std. 1.1), indicating that on average 3--4 waypoints (0.3--0.4s of predicted trajectory) are discarded at each splice due to the accumulated latency. All waypoints before $i^{*}$ are discarded; execution resumes from the spatially aligned point.

\section{Experiments}
\label{sec:experiments}

The evaluation answers three questions: whether the data pipeline produces trajectories of demonstration quality, how Mobile UMI compares against representative policy baselines under matched settings, and how each of the two proposed components contributes to the overall gain.

\subsection{Experimental Setup}
\label{sec:setup}

The workspace consists of two tables in an L-shaped layout (Table~A: 100$\times$40$\times$100cm; Table~B: 80$\times$40cm, adjustable height). The robot operates within a 3$\times$3m floor area. Hardware specifications are listed in Section~\ref{sec:hardware}.

\textbf{Training details.}
All policies are trained on a single RTX~4070 (12GB) GPU using AdamW with an initial learning rate of $1\times10^{-4}$ and weight decay of $10^{-6}$, linearly warmed up over 500 steps and then cosine-annealed. Batch size is 64; training runs for 20 epochs per task. The diffusion schedule uses $K{=}100$ steps with cosine-squared $\bar{\alpha}_k$, reduced to 10 DDIM steps at inference. A separate network is trained for each task. ACT follows the official release (4-layer transformer encoder-decoder, chunk size 16, KL weight 10). Diffusion Policy reuses the visual encoder, FiLM conditioning, and 1D U-Net backbone of Mobile UMI.

Four tasks of increasing difficulty are evaluated:
\begin{enumerate}
\item \textbf{Tea Bag Placement}: navigate to a table, grasp a tea bag, transport it, place it into a cup.
\item \textbf{Turn off the light}: navigate to a wall-mounted light switch and flip it off.
\item \textbf{Book placement}: pick up books and place them at a target location.
\item \textbf{Household Manipulation}: a composite household-style episode combining navigation with multi-step object handling.
\end{enumerate}

Demonstration counts per task are: Tea Bag Placement 50, Turn off the light 100, Book placement 100, and Household Manipulation 200, all collected by a single operator following the protocol in Section~\ref{sec:assembly}. All demonstrations are recorded in the same laboratory environment under consistent overhead lighting (500 lux). Object poses are manually reset to nominal positions before each demonstration, with natural variation from human placement imprecision (typically $\pm$2cm in translation, $\pm$5$^{\circ}$ in orientation). For deployment evaluation, we run 100 independent trials per task. Each trial starts from a randomized initial base pose within a 10cm radius and $\pm$15$^{\circ}$ heading variation of a nominal position. A trial is marked successful only if the complete interaction sequence finishes without human intervention within a 120s time limit.

\subsection{Data Collection Pipeline Validation}
\label{sec:pipeline_val}

\begin{figure*}[!t]
\centering
\includegraphics[width=\textwidth]{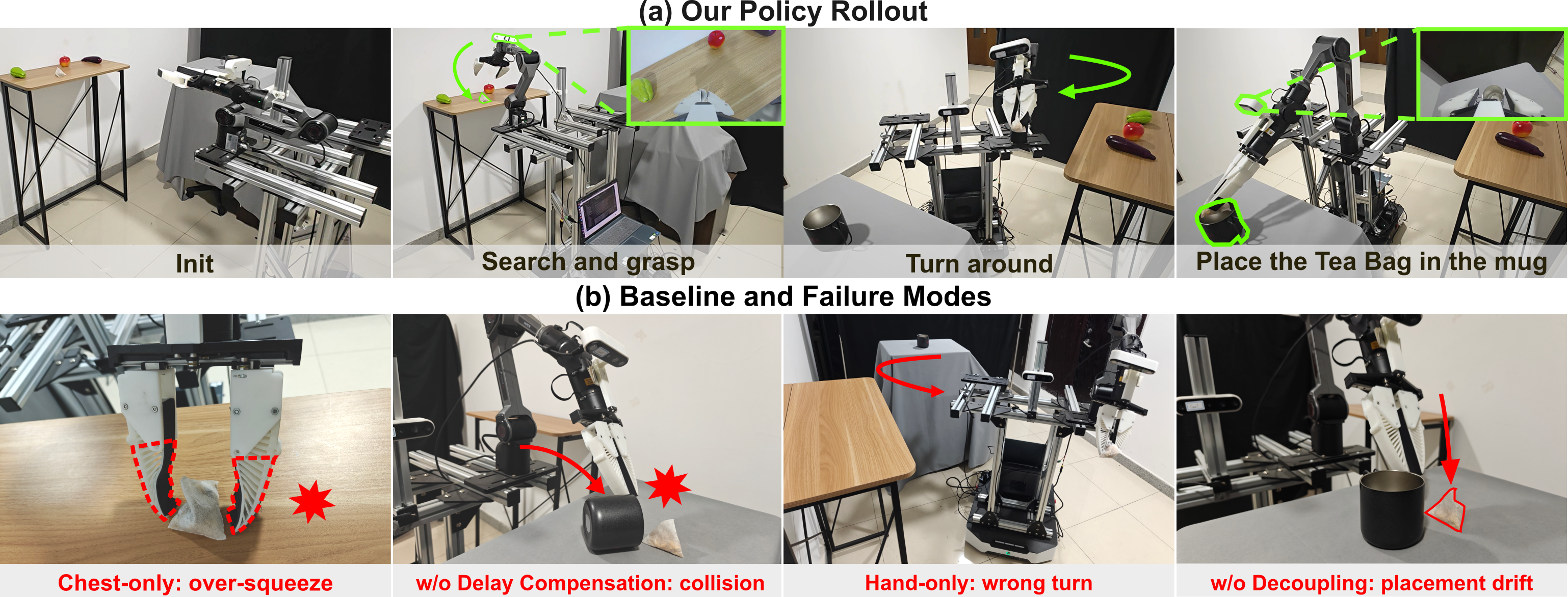}
\caption{Tea Bag Placement: qualitative rollout and representative failure modes.
Top row shows a successful sequence (init, search-and-grasp, turn-around, place into mug).
Bottom row highlights typical failures under module removal settings.}
\label{fig:tasks_tea}
\end{figure*}

\textbf{VIO closed-loop test.}
The demonstration-end camera was rigidly mounted on the Piper arm end-effector, which executed a 120s trajectory spanning approximately 10m. After returning to the start pose, the position closure error measured 14.1mm (0.14\% of path length), with attitude drift below 1.63$^{\circ}$ in yaw.

\textbf{Anchoring reprojection.}
Over five repeated 200s collection sessions, the anchoring residual yielded a position RMS of 12.1mm and a rotation RMS of 0.86$^{\circ}$.

\textbf{Open-loop replay.}
Twenty open-loop replays of a 2m trajectory produced end-effector landing scatter with X-axis RMS below 2cm and Y-axis RMS of approximately 4cm.

\subsection{Comparison Studies}
\label{sec:task_results}

Table~\ref{tab:results} compares Mobile UMI against ACT and Diffusion Policy baselines. To isolate the representation from the policy class, each baseline is trained with both the global hand pose and the chest-relative label of Section~\ref{sec:decouple}; all variants share the same demonstrations, observations, and training schedule. Success rates are reported as percentages over 100 independent trials per task.

\begin{table*}[!t]
\renewcommand{\arraystretch}{1.2}
\caption{Task Success Rates (\%) over 100 Trials: Baseline Comparison and Ablation Study}
\label{tab:results}
\centering
\small
\begin{tabular}{lcccccc}
\toprule
\multirow{2}{*}{\textbf{Method}} &
\multicolumn{4}{c}{\textbf{Task Success Rate (\%)}} &
\multirow{2}{*}{\textbf{Avg.~(\%)}} &
\multirow{2}{*}{\textbf{Avg.~Time (s)}} \\
\cmidrule(lr){2-5}
 & \makecell{\textbf{Tea Bag}\\\textbf{Placement}} & \makecell{\textbf{Turn off the}\\\textbf{light}} & \makecell{\textbf{Book}\\\textbf{placement}} & \makecell{\textbf{Household}\\\textbf{Manipulation}} & & \\
\midrule
ACT (global label)                          & 55 & 45 & 38 & 32 & 42.5 & 84.2 \\
ACT (chest-relative label)                  & 68 & 58 & 45 & 35 & 51.5 & 78.5 \\
Diffusion Policy (global label)             & 65 & 56 & 50 & 39 & 52.5 & 78.6 \\
Diffusion Policy (chest-relative label)     & 80 & 72 & 63 & 52 & 66.8 & 68.3 \\
\textbf{Mobile UMI (Ours)}                  & \textbf{95} & \textbf{90} & \textbf{80} & \textbf{70} & \textbf{83.8} & \textbf{59.7} \\
\midrule
w/o Decoupled Representation                & 30 & 20 & 15 & 10 & 18.8 & N/A \\
w/o Latency Compensation                    & 75 & 65 & 55 & 45 & 60.0 & 72.3 \\
\bottomrule
\end{tabular}
\end{table*}

The seven rows differ in policy class, action label, and state matching. \emph{ACT (global / chest-relative)} reproduces the Action Chunking Transformer~\cite{b27} with world-frame and decoupled labels. The ACT baseline was adapted to predict identical relative increments under the same receding-horizon execution. \emph{Diffusion Policy (global / chest-relative)} reproduces the original Diffusion Policy~\cite{b2} under both labels. \emph{Mobile UMI (Ours)} combines the chest-relative label with state matching. The ablation rows isolate each component: \emph{w/o Decoupled Representation} keeps state matching but uses the global label; \emph{w/o Latency Compensation} keeps the decoupled label but disables state matching.

Mobile UMI achieves 83.8\% average success rate, outperforming Diffusion Policy with chest-relative labels (66.8\%) by 17.0 percentage points, with largest gains on Book placement (80\% vs. 63\%) and Household Manipulation (70\% vs. 52\%). Switching from global to chest-relative labels improves both baselines: ACT from 42.5\% to 51.5\%, Diffusion Policy from 52.5\% to 66.8\%.

\subsection{Ablation Studies}
\label{sec:ablation}

\textbf{Without decoupled representation.}
Removing kinematic decoupling causes catastrophic failure (18.8\% success rate). Treating the hand pose in the global frame conflates intentional manipulation with walking-induced motion, making it impossible to learn a coherent manipulation policy.

\textbf{Without delay compensation.}
Disabling state matching reduces success rate from 83.8\% to 60.0\%. The controller executes stale action chunks that no longer match the robot's current pose, causing trajectory rollbacks. The impact is most severe on Book placement (55\% vs. 80\%) and Household Manipulation (45\% vs. 70\%).

Together, these ablations show that kinematic decoupling provides the dominant contribution, while latency compensation is essential for smooth execution.

\section{Conclusion}
This paper presented Mobile UMI, a framework that enables scalable demonstration collection and robust policy deployment for mobile manipulation. A dual-camera portable system records chest-centric navigation and wrist-centric manipulation views, from which a spatial anchoring pipeline extracts decoupled $\mathrm{SE}(2)$ base trajectories and relative $\mathrm{SE}(3)$ hand trajectories. A cross-view conditional diffusion policy trained on these representations generates composite actions that respect the kinematic constraints of both the base and the arm. An asynchronous receding-horizon executor with online state matching bridges the gap between slow generative inference and the high-frequency low-level controller, ensuring smooth execution on a continuously moving platform. Real-world experiments across four household tasks validated the effectiveness of each component, demonstrating substantial improvements over existing baselines.

A limitation of kinematic decoupling is the inability to capture whole-body dynamic force exchanges. While effective for sequential "carry-and-move" tasks, heavy payload manipulation where the base must counteract arm wrenches may yield sub-optimal compliance. Future work will explore dynamic force-closure integration and extend the framework to non-holonomic platforms with richer motion primitives.

\suppressfloats[t]
\begin{figure*}[!t]
\centering
\includegraphics[width=0.95\textwidth]{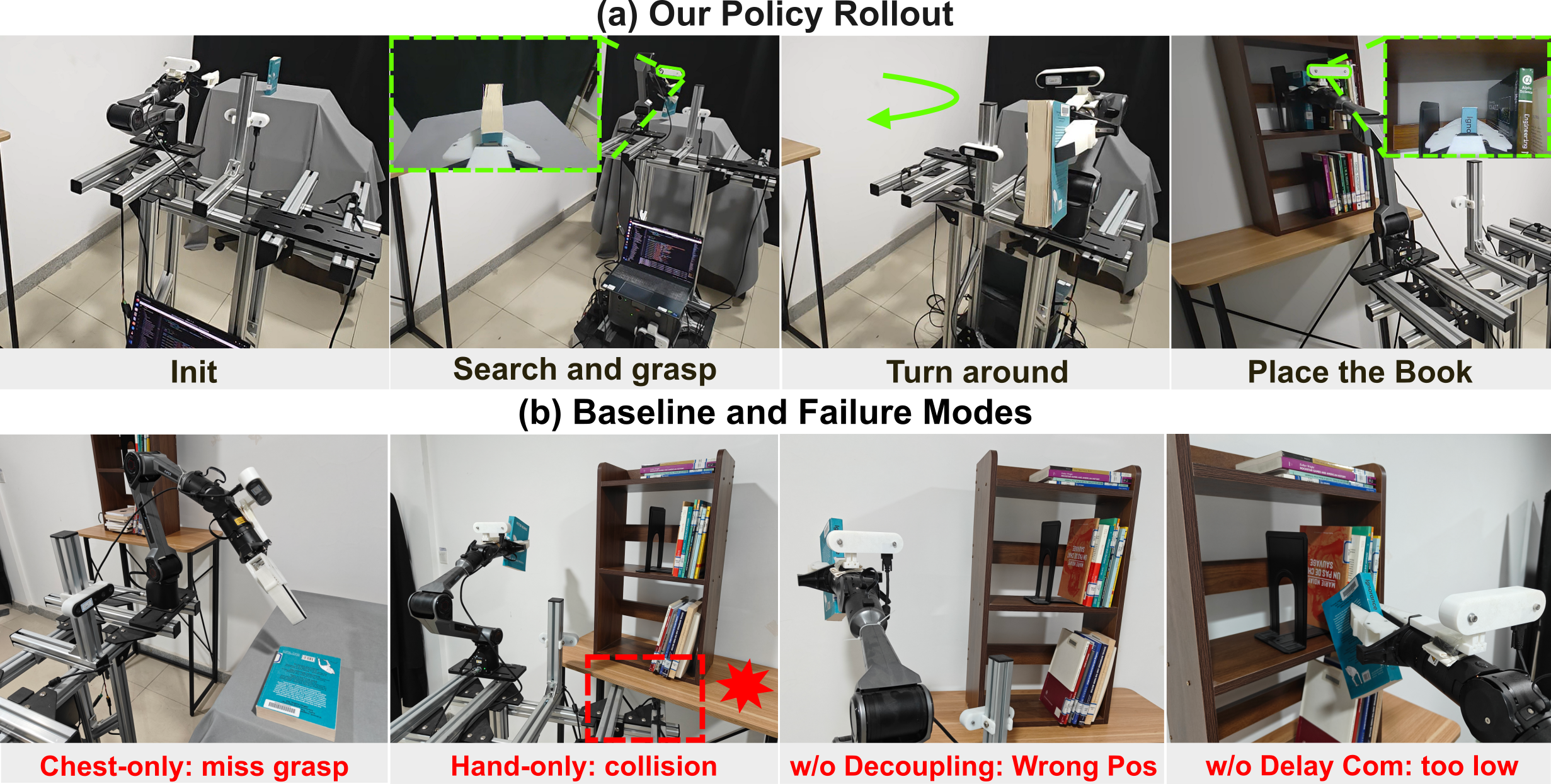}
\caption{Book placement: rollout and baseline failure modes under different module ablations.}
\label{fig:tasks_book}
\end{figure*}

\begin{figure}[!t]
\centering
\includegraphics[width=\columnwidth]{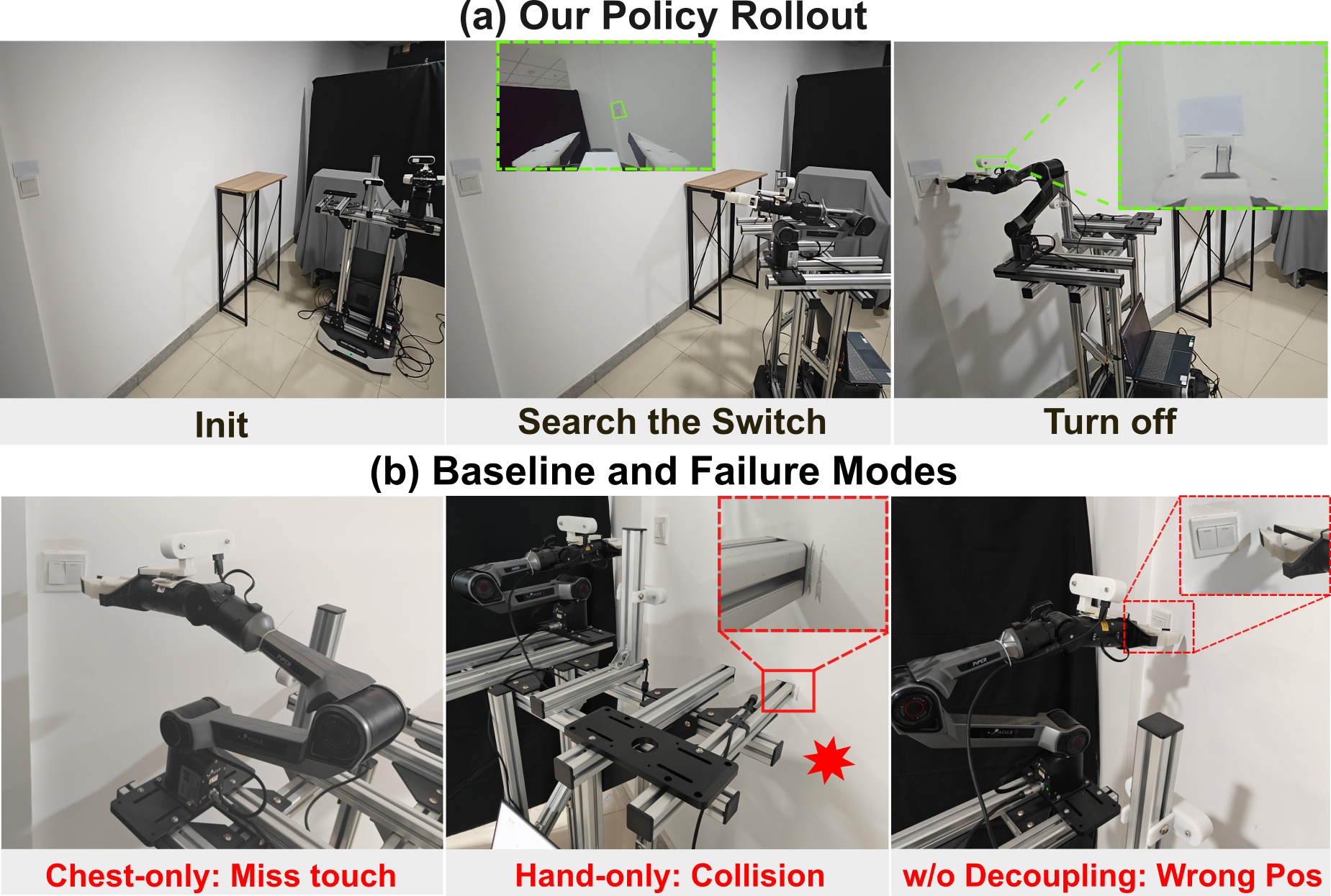}
\caption{Turn off the light: rollout and representative failure modes (miss touch, collision, wrong pose).}
\label{fig:tasks_light}
\end{figure}


\end{document}